\documentclass{article}
\usepackage{graphicx} 

\usepackage[preprint]{corl_2026} 

\usepackage[T1]{fontenc}
\usepackage{booktabs}
\usepackage{multirow}
\usepackage{amsmath,amssymb}
\usepackage{subcaption}
\usepackage{float}
\newcommand{\pp}{\,\mathrm{pp}}
\newcommand{\VO}{\textsc{V}}
\newcommand{\VT}{\textsc{V+T}}


\title{Tactile Sensors Are an Effective Substitute for Wrist Cameras in Contact-Rich or Occluded Manipulation
}

\author{
  Aijia Yang \quad Arsh Tangri \quad
  Sergey Zakharov \quad Krishna Murthy Jatavallabhula \\
  Toyota Research Institute \\
  \texttt{any015@ucsd.edu} \\
  \url{https://annieyaj.github.io/tactile-vs-vision/}
}

\begin{document}
\maketitle
\hypersetup{
  pdfauthor={Aijia Yang, Arsh Tangri,
    Sergey Zakharov, Krishna Murthy Jatavallabhula},
  pdftitle={Tactile Sensors Are an Effective Substitute
    for Wrist Cameras in Contact-Rich or Occluded Manipulation}
}


\begin{abstract}

Learned manipulation policies often rely on wrist-mounted cameras for close-range visual feedback that complements external views of the workspace. However, contact-rich manipulation requires establishing and maintaining contacts that can remain hidden even from wrist cameras. 
Tactile sensors mounted on fingers can reliably and directly measure contact, without needing visual access.
In this work, we show that this feedback enables policies to recover the performance advantage of wrist cameras using scene and egocentric vision together with touch. 
Our approach incorporates two low-cost tactile arrays into a visuotactile ACT policy, without wrist images during training or deployment. 
We evaluate six real-world task families spanning placement, insertion, occluded retrieval, and sustained-contact tool use. On three contact-rich tasks, we independently vary wrist-camera and tactile availability while keeping the demonstrations, policy architecture, and optimization settings fixed. Adding wrist cameras improves success by 18.3--28.3 percentage points over vision without wrist images. Adding tactile sensing instead improves success by 31.7--60.0 percentage points and yields higher observed success than vision-only policies with wrist cameras on all three tasks, with significant advantages on two under Holm-corrected tests.
These results support finger-mounted tactile sensing as an effective alternative to wrist cameras in the evaluated contact-rich settings.

    
\end{abstract}



\section{Introduction}
	
Learned manipulation policies rely on external cameras for scene context and wrist-mounted cameras for close-range observations of the gripper and manipulated object. However, during manipulation, the fingers can occlude grasped objects, and whether contact is maintained may be difficult to observe even with wrist cameras. Tactile sensors provide direct measurements at the finger–object interface, supplying feedback without requiring visual access to the contact. This raises two research questions: (a) \textit{Can tactile sensing substitute for, and improve on, wrist cameras?} and (b) \textit{When?} We investigate these questions by comparing visuotactile and vision-only policies while controlling wrist-camera availability.

Using the Raiden toolkit~\citep{raiden}, we train and deploy Action Chunking with Transformers (ACT)~\citep{act} policies on a YAM robot arm equipped with two low-cost FlexiTac~\citep{flexitac} tactile sensor arrays. Our primary configuration combines scene and egocentric images with tactile observations, without wrist images during training or deployment. 
We evaluate multiple tasks spanning placement, fine alignment, occluded retrieval, and sustained contact. 
We compare vision-only and visuotactile policies while varying the availability of wrist cameras, under the same trajectory data distribution. Our contribution is a controlled empirical study of when finger-mounted tactile input can substitute for wrist images in an otherwise fixed ACT pipeline.

We observe substantial success gains from adding wrist cameras for vision-only policies. However, we also find that adding tactile sensing results in similar (or better) performance, where vision+tactile (\VT) policies without wrist cameras achieve higher observed success than vision-only (\VO) policies with wrist cameras on contact-rich tasks. 
These results support tactile sensing as an effective substitute for wrist cameras in the evaluated contact-rich or occluded manipulations.
    

\section{Related Work}
\label{sec:related_work}

    Low-cost tactile sensors such as FlexiTac~\citep{flexitac} and eFlesh~\citep{eflesh} have made contact sensing increasingly accessible. TacO~\citep{taco} shows that tactile benefits depend on both task and sensor modality, rather than providing uniform improvements. Earlier approaches such as Visuo-Tactile Transformers~\citep{visuo22} and 3D-ViTac~\citep{3dvitac} demonstrate the benefits of learned visual-tactile representations for manipulation. We use FlexiTac to study how the benefit of tactile input changes with visual observations available to a learned manipulation policy. 

    Closest to our setting, LeFlexiTac~\citep{leflexitac} integrates FlexiTac into several manipulation policies and reports particularly large gains for occluded retrieval. VT-Refine~\citep{vtrefine} and T-Rex~\citep{t-rex} also demonstrate the value of tactile feedback for assembly and contact-rich dexterity. Whereas these works establish that touch can help, we hold the sensor type and policy class fixed and vary tactile input and wrist-vision availability, examining whether touch provides benefits beyond wrist vision and whether it can substitute for wrist vision when scene and egocentric observations remain available. 
        	

\section{Methodology}
\label{sec:methodology}

    \begin{figure}[t]
    \centering

        \begin{subfigure}[t]{0.3\linewidth}
            \centering
            \includegraphics[height=3cm, keepaspectratio]{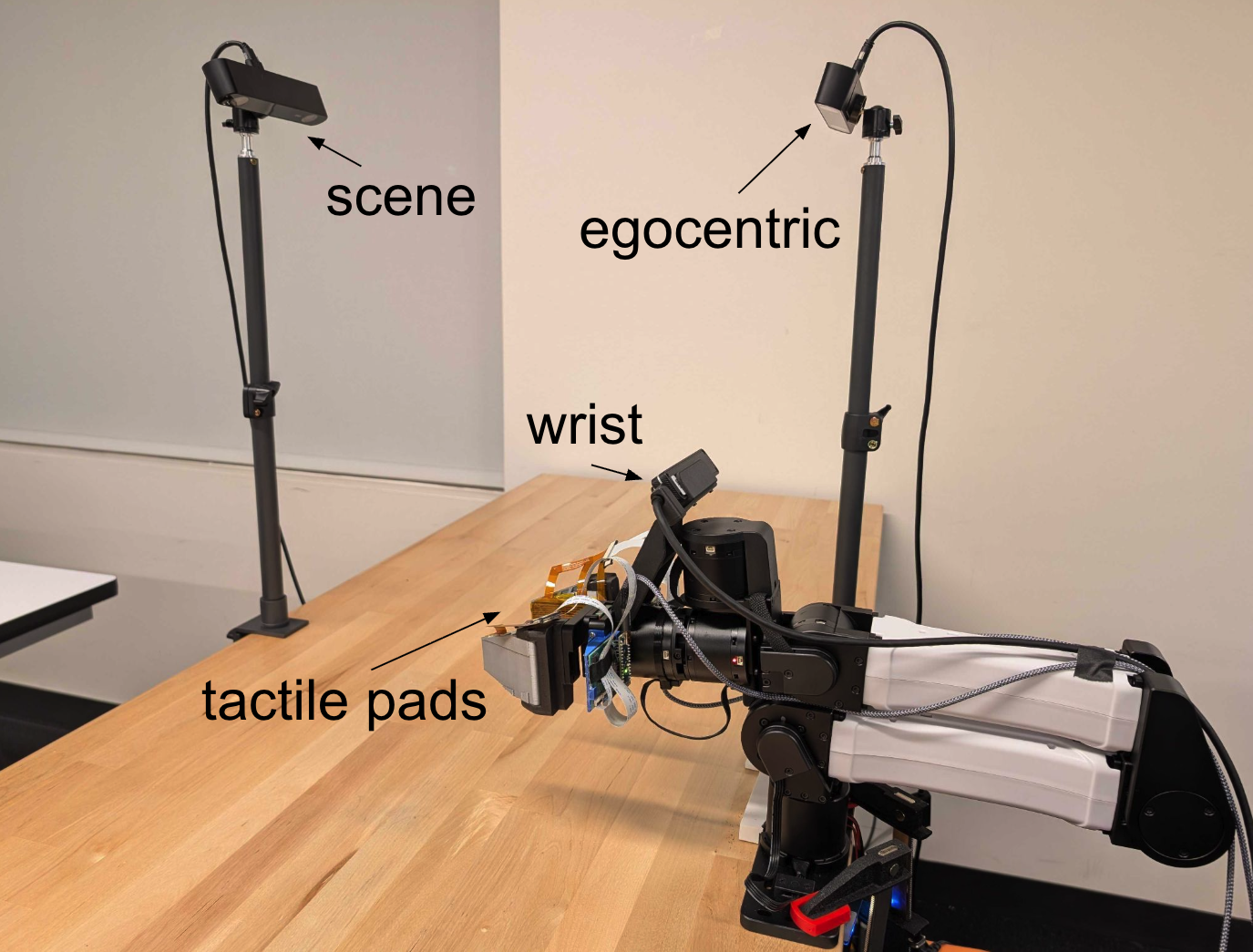}
            \caption{station setup}
            \label{fig:a}
        \end{subfigure}
        \hspace{0.003\linewidth}
        \begin{subfigure}[t]{0.3\linewidth}
            \centering
            \includegraphics[height=3cm, keepaspectratio]{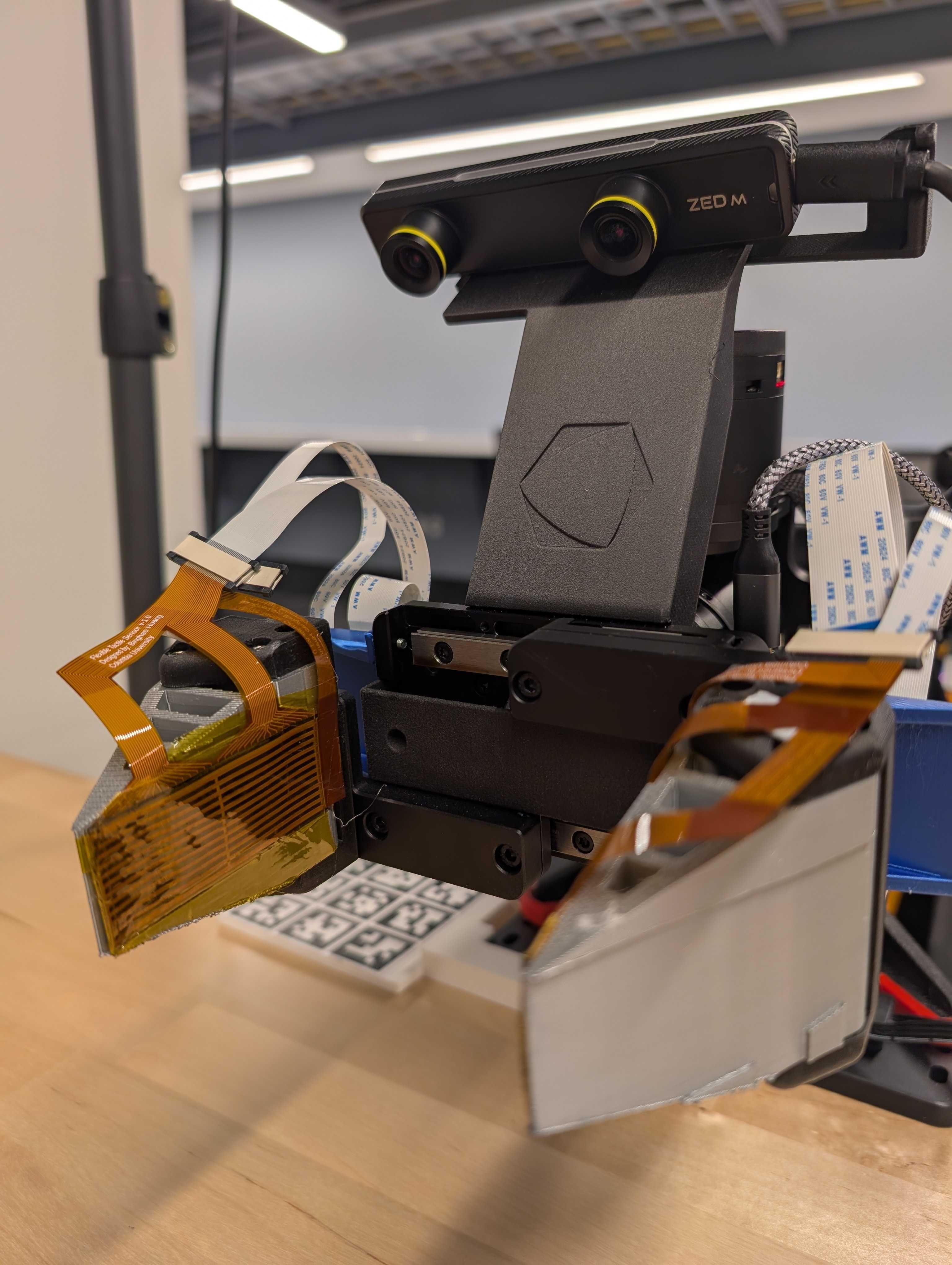}
            \caption{tactile sensor}
            \label{fig:b}
        \end{subfigure}
        \hspace{0.003\linewidth}
        \begin{subfigure}[t]{0.35\linewidth}
            \centering
            \includegraphics[height=3cm, keepaspectratio]{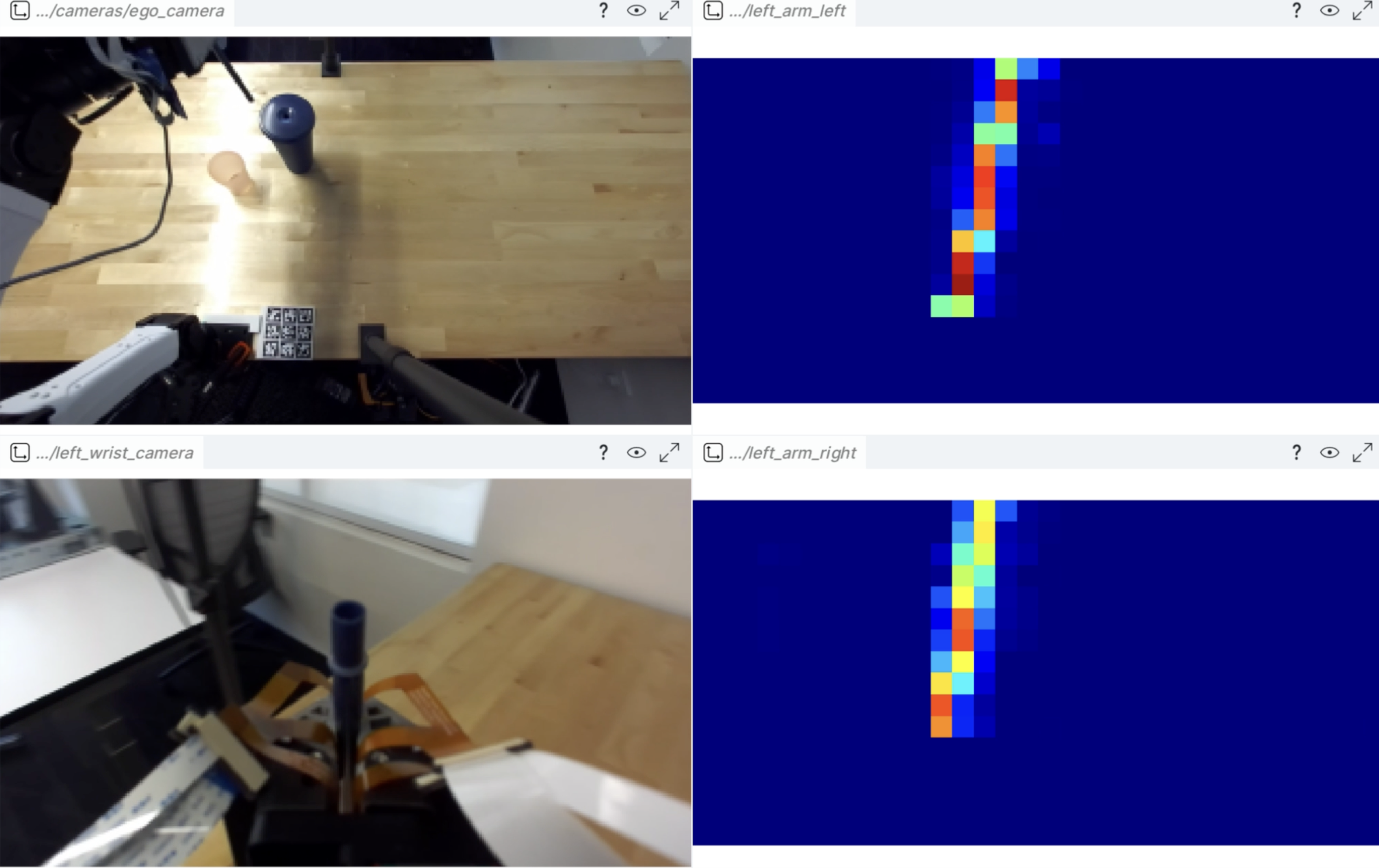}
            \caption{T3 tactile heatmap}
            \label{fig:c}
        \end{subfigure}
    
        \vspace{2mm}
        
        \begin{subfigure}[t]{0.158\linewidth}
            \centering
            \includegraphics[width=\linewidth]{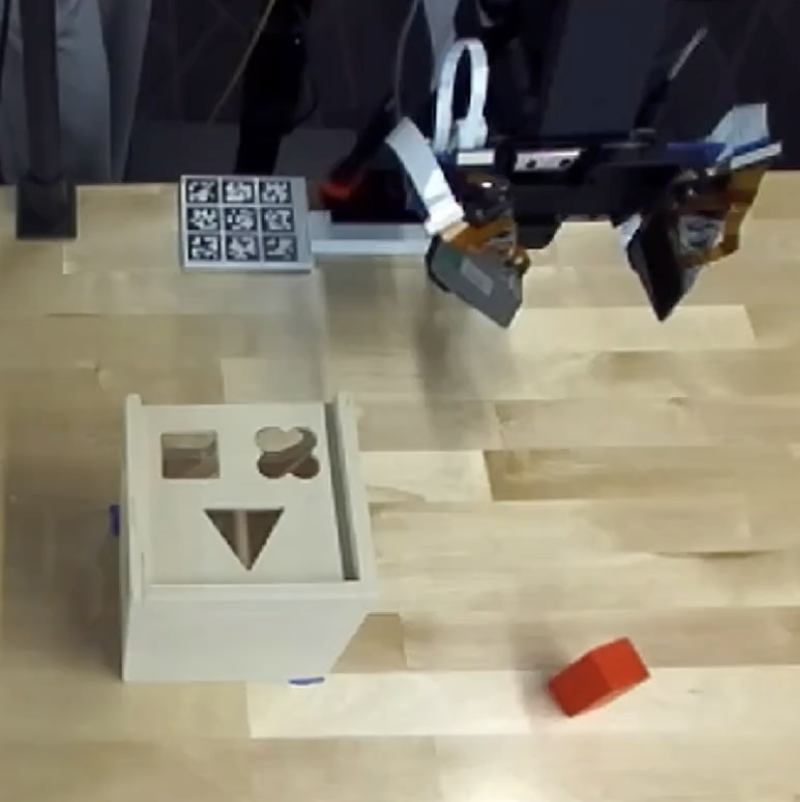}
            \caption{}
            \label{fig:d}
        \end{subfigure}
        \hfill
        \begin{subfigure}[t]{0.158\linewidth}
            \centering
            \includegraphics[width=\linewidth]{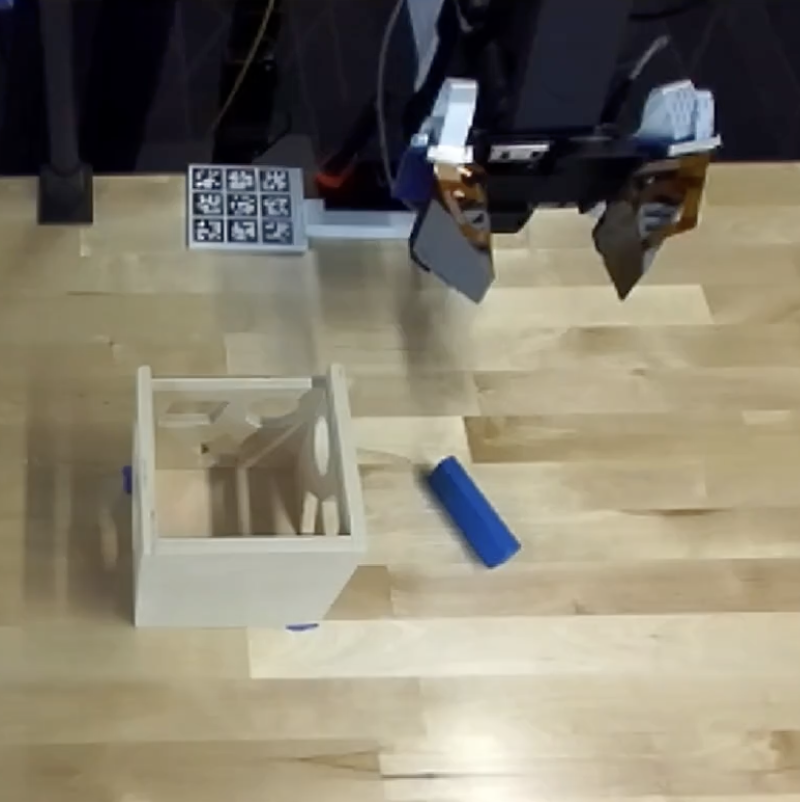}
            \caption{}
            \label{fig:e}
        \end{subfigure}
        \hfill
        \begin{subfigure}[t]{0.158\linewidth}
            \centering
            \includegraphics[width=\linewidth]{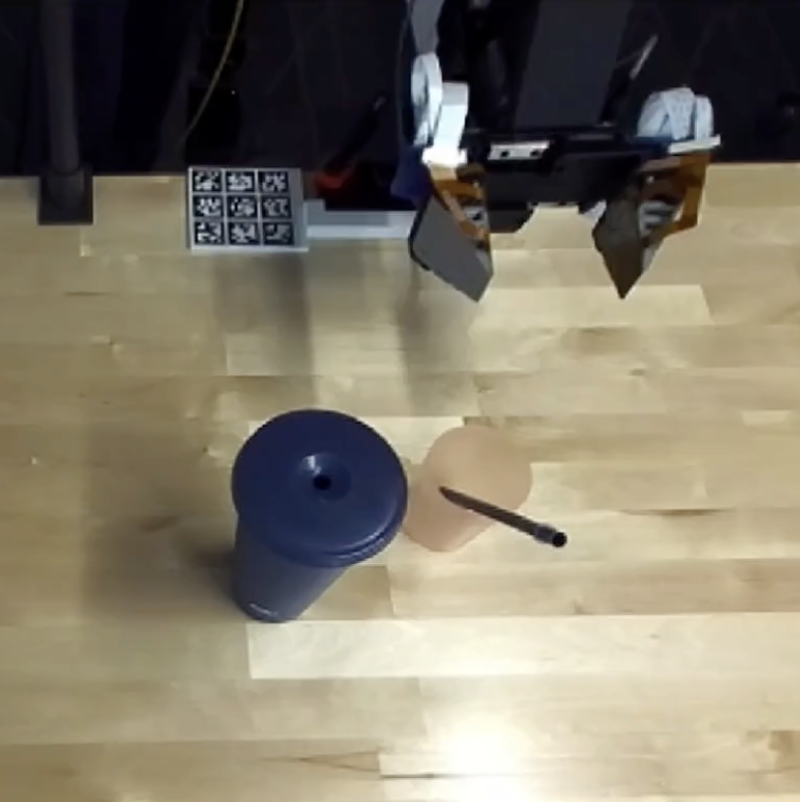}
            \caption{}
            \label{fig:f}
        \end{subfigure}
        \hfill
        \begin{subfigure}[t]{0.158\linewidth}
            \centering
            \includegraphics[width=\linewidth]{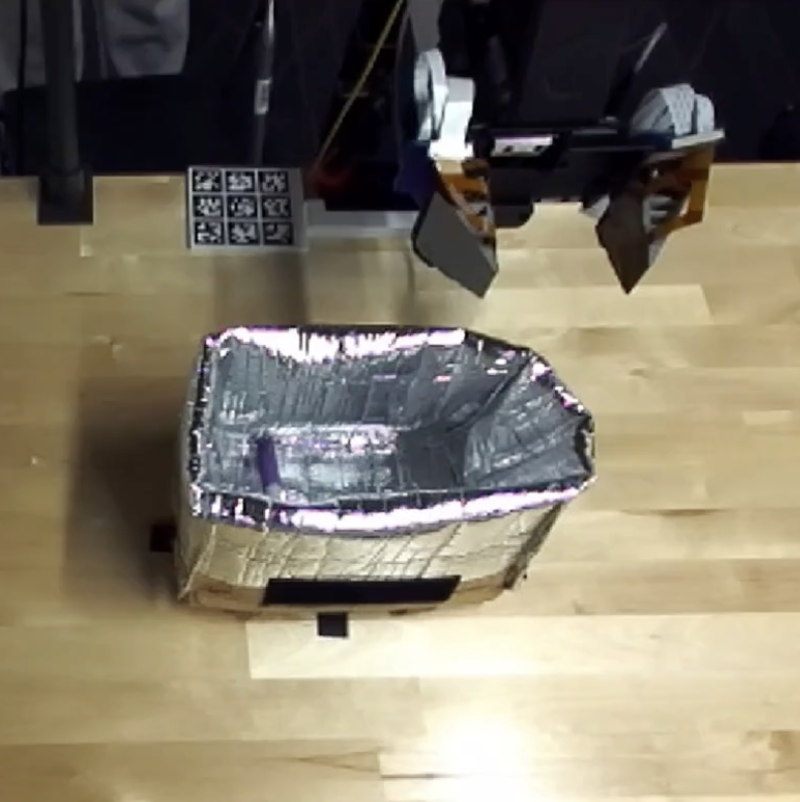}
            \caption{}
            \label{fig:g}
        \end{subfigure}
        \hfill
        \begin{subfigure}[t]{0.158\linewidth}
            \centering
            \includegraphics[width=\linewidth]{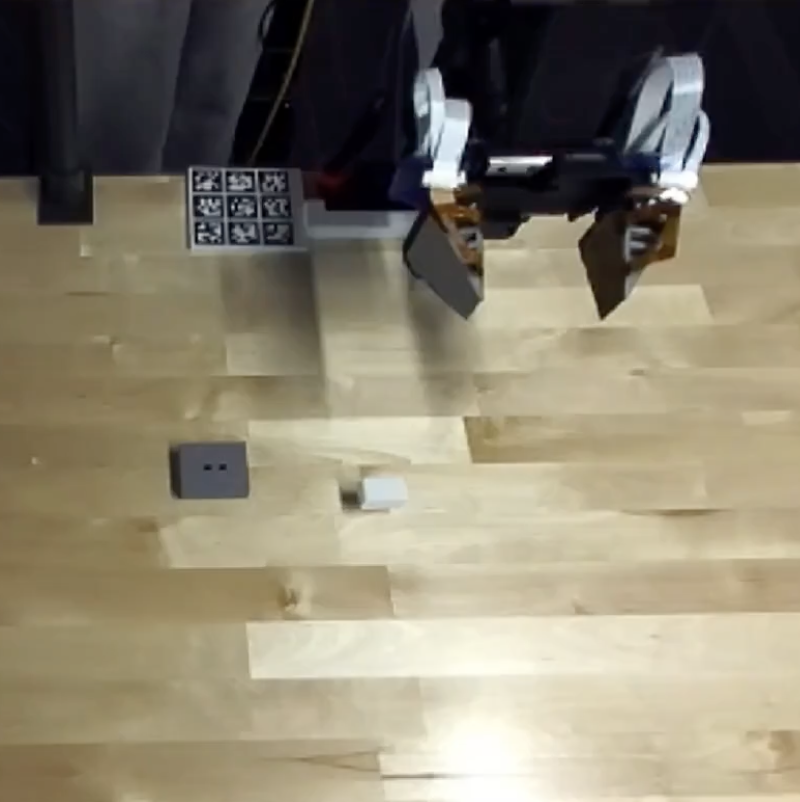}
            \caption{}
            \label{fig:h}
        \end{subfigure}
        \hfill
        \begin{subfigure}[t]{0.158\linewidth}
            \centering
            \includegraphics[width=\linewidth]{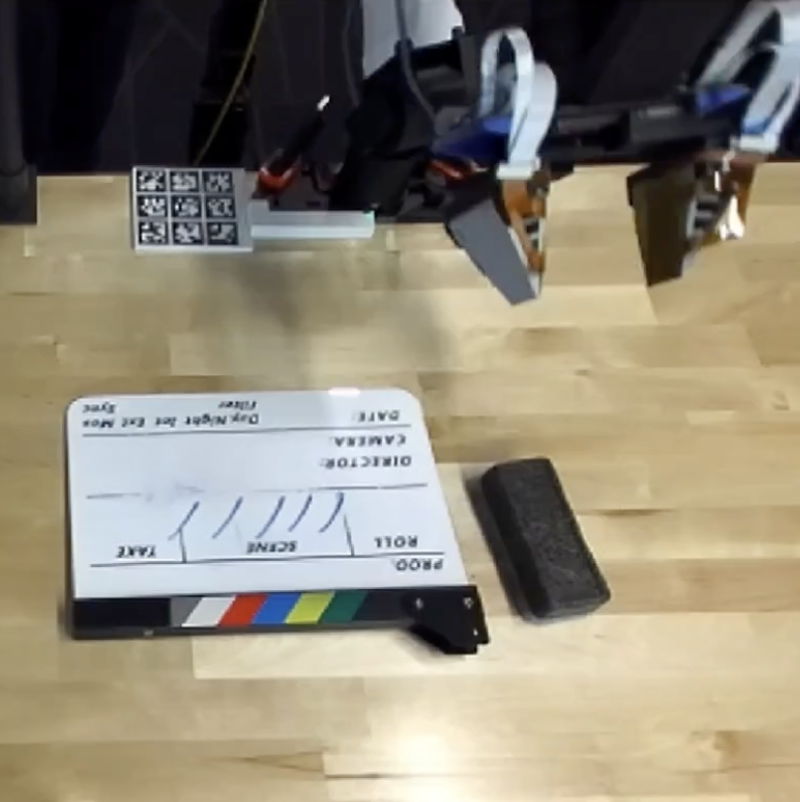}
            \caption{}
            \label{fig:i}
        \end{subfigure}
        \caption{\textbf{System and evaluation tasks.}
        Two FlexiTac pads are mounted on the inner gripper surfaces and recorded alongside scene, egocentric, and wrist RGB observations. Panels~\subref{fig:d}--\subref{fig:i} show tasks T1--T6, spanning placement, fine alignment, occluded retrieval, and sustained contact.}
        \label{fig:system}
    \end{figure}

    Experiments use a YAM robotic platform equipped with scene, egocentric, and wrist RGB cameras (Fig.~\ref{fig:a}). Throughout this paper, without wrist vision means that scene and egocentric images are retained and only wrist images are omitted. Two FlexiTac~\citep{flexitac} piezoresistive arrays are mounted on the inner surfaces of the TPU compliant gripper fingers (Fig.~\ref{fig:b}, Appendix \ref{app:hardware}). Tactile sensing is recorded at approximately $99$~Hz and RGB observations at approximately $30$~Hz using a shared clock.

    During dataset construction, tactile measurements are aligned to each policy timestep by nearest timestamp. The full-rate tactile stream is retained in recordings rather than down-sampled to camera frequency, allowing future use of high-frequency contact information. 

    All experiments use ACT (Action Chunking with Transformers)~\citep{act}, predicting 19-step chunks of absolute 7-DoF left-arm joint position commands (Training details in Appendix \ref{app:training}). We compare \textbf{vision-only (\VO)}, which receives RGB observations, against \textbf{vision+tactile (\VT)}, which additionally receives two $32{\times}16$ tactile maps. Each grayscale tactile map is replicated to three channels, resized to $224{\times}224$, and processed by the same shared ImageNet-pretrained ResNet-18~\citep{resnet} used for camera views. Tactile sensing therefore enters ACT as two additional image-like observations. The policy class, demonstration data, action representation, optimization settings, and deployment procedures are otherwise fixed between observation conditions. Checkpoint selection is described separately in Appendix~\ref{app:training}.

    We evaluate tight-clearance block placement (T1), rigid-block and deformable-sponge placement (T2a/b), straw insertion (T3), opaque-bag marker retrieval (T4), plug insertion (T5), and whiteboard wiping (T6). T1--T2 are largely visually observable placement tasks; T3 and T5 require precise alignment; T4 involves retrieving an object from inside an opaque bag; and T6 additionally requires maintaining contact during wiping.

    For each task, \VO\ and \VT\ are evaluated from matched initial-condition sets using fixed success criteria and rollout counts. For T4--T6, both policies are trained and evaluated with and without wrist images. The evaluation comprises 1,000 rollouts across 20 task-and-observation conditions, counting T2a and T2b separately. Dataset construction, episode-level splits, and evaluation allocations are described in Appendix~\ref{app:training}. 

\section{Experimental Results}
\label{sec:result}

    \begin{table}[t]
      \centering
      \small
      \caption{\textbf{Vision-only vs.\ vision+tactile success.}
      Each policy entry reports successes/rollouts (success rate). T4--T6 are evaluated both with and without wrist images. $\Delta$ is the observed \VT-minus-\VO\ difference in percentage points.}
      \label{tab:results}
      \begin{tabular}{@{}llcccc@{}}
        \toprule
         & Task & Wrist & \VO & \VT & $\Delta$ \\
        \midrule
        T1 & Tight-clearance block & on & 15/30 (50.0\%) & 19/30 (63.3\%) & $+13.3\pp$ \\
        T2a & Rigid block & on & 15/25 (60.0\%) & 15/25 (60.0\%) & $0.0\pp$ \\
        T2b & Sponge & on & 19/25 (76.0\%) & 22/25 (88.0\%) & $+12.0\pp$ \\
        T3 & Straw insertion & on & 26/60 (43.3\%) & 39/60 (65.0\%) & $+21.7\pp$ \\
        \midrule
        \multirow{2}{*}{T4} & \multirow{2}{*}{Opaque-bag retrieval}
          & off & 13/60 (21.7\%) & 49/60 (81.7\%) & $+60.0\pp$ \\
         & & on & 30/60 (50.0\%) & 43/60 (71.7\%) & $+21.7\pp$ \\
        \multirow{2}{*}{T5} & \multirow{2}{*}{Plug insertion}
          & off & 9/60 (15.0\%) & 28/60 (46.7\%) & $+31.7\pp$ \\
         & & on & 20/60 (33.3\%) & 30/60 (50.0\%) & $+16.7\pp$ \\
        \multirow{2}{*}{T6} & \multirow{2}{*}{Whiteboard wiping}
          & off & 13/60 (21.7\%) & 43/60 (71.7\%) & $+50.0\pp$ \\
         & & on & 28/60 (46.7\%) & 39/60 (65.0\%) & $+18.3\pp$ \\
        \bottomrule
      \end{tabular}
    \end{table}

    Table~\ref{tab:results} reports observed success rates. Posterior summaries, tactile-gain intervals, and Holm-corrected tests are provided in Appendix~\ref{app:posteriors}. Counts by initial condition are in Appendix~\ref{app:counts}.
    
    \paragraph{Smaller gains on placement tasks. }
    Tactile input leaves the observed rigid-block success unchanged at $60.0\%$ and improves tight-clearance and sponge placement by $13.3\pp$ and $12.0\pp$, respectively. None of these differences are statistically significant under the corrected tests.
    
    \paragraph{Touch helps with fine alignment.}
    Straw insertion success increases from $43.3\%$ to $65.0\%$ ($+21.7\pp$). This improvement occurs despite the target being visible, showing that tactile benefits are not limited to fully occluded manipulation.

    \paragraph{Largest gains without wrist vision.}
    Without wrist vision, tactile input increases success by $60.0\pp$ for opaque-bag retrieval, $31.7\pp$ for plug insertion, and $50.0\pp$ for whiteboard wiping. These tasks involve occluded grasping, fine insertion, and sustained contact, respectively.

    \subsection{Wrist-Camera Control Comparison}
    \label{sec:wrist}
    Adding wrist vision improves \VO\ success by $28.3\pp$, $18.3\pp$, and $25.0\pp$ on T4--T6, respectively. The corresponding tactile gains shrink to $21.7\pp$, $16.7\pp$, and $18.3\pp$.

    Adding wrist images does not consistently improve \VT. Observed success changes from $81.7\%$ to $71.7\%$ on T4, from $46.7\%$ to $50.0\%$ on T5, and from $71.7\%$ to $65.0\%$ on T6. The wrist-on and wrist-off \VT\ conditions are not significantly different on T4 or T6 under the Holm-corrected tests (Appendix~\ref{app:posteriors}), although this does not establish equivalence. Smaller marginal tactile gains are consistent with overlapping task-relevant information, but information overlap alone does not explain the lower absolute point estimates. Finite evaluation samples, training and checkpoint variability, and differences in how the learned policies use the available observations are possible explanations; the present experiment does not distinguish among them.

    For T4, the lower \VT\ wrist-on success is concentrated in far-right trials ($9/20$ vs.\ $15/20$ without wrist images; Table~\ref{tab:t4_positions}). This post hoc comparison does not establish causality.
    
    \subsection{Can Touch Substitute for Wrist Vision?}
    \label{sec:substitute}
    
    We directly compare \VT\ without wrist images against \VO\ with wrist images. Observed success is higher with touch on all three contact-rich tasks: $81.7\%$ vs.\ $50.0\%$ for opaque-bag retrieval, $46.7\%$ vs.\ $33.3\%$ for plug insertion, and $71.7\%$ vs.\ $46.7\%$ for whiteboard wiping. The differences are statistically significant for T4 and T6 under the Holm-corrected tests.

    These results support touch as an alternative to wrist vision on the evaluated contact-rich tasks, not as a replacement for vision overall: both conditions retain scene and egocentric images.

    \paragraph{Stagewise outcomes in wiping.} In T6, tactile input increases observed eraser-pickup success from $23/60$ to $48/60$ without wrist images and from $31/60$ to $42/60$ with wrist images. Task success given pickup increases from $13/23$ to $43/48$ without wrist images, but is similar with wrist images ($28/31$ vs.\ $39/42$).
    
    Comparing the two \VT\ conditions, observed pickup success is $48/60$ without wrist images and $42/60$ with wrist images, while task success given pickup is $43/48$ ($89.6\%$) and $39/42$ ($92.9\%$), respectively. The pickup decrease is concentrated in the bottom initial position, with $15/20$ successful pickups without wrist images and $9/20$ with wrist images (Table~\ref{tab:t6_positions}). Thus, lower overall wrist-on success coincides with fewer successful pickups, while observed completion among successful pickups remains high. These comparisons are descriptive: the successful-pickup subsets differ across policies, and the position-level analysis does not establish a causal mechanism.
    

\section{Discussion and Limitations}
\label{sec:discussion}

    The benefit of touch depends strongly on the task and on how much close-range visual information is available. Observed gains are smaller on placement tasks and largest on T4--T6 without wrist images. Adding wrist vision narrows the tactile advantage but does not consistently improve \VT; the cause of the lower wrist-on success rates remains unresolved. Without wrist images, \VT\ achieves higher observed success than \VO\ with wrist images on all three contact-rich tasks, with significant advantages on T4 and T6. In these tasks, touch can recover---and in some cases exceed---the benefit of adding wrist vision, while scene and egocentric vision remain available.


    Our study is limited to one robot platform, one tactile sensor type, one policy-learning method, and one training seed. Low-cost tactile pads also exhibit baseline variation after replacement, motivating improved normalization and evaluation across additional policies and sensors. Overall, the results suggest that vision guides the robot toward contact, while touch resolves interaction states that cameras cannot reliably observe. 


\clearpage


\bibliography{references}  

\clearpage
\appendix

\section{Hardware Designs}
\label{app:hardware}

    \begin{figure}[h]
    \centering
        \begin{subfigure}[t]{0.28\linewidth}
            \centering
            \includegraphics[width=\linewidth, height=6cm, keepaspectratio]{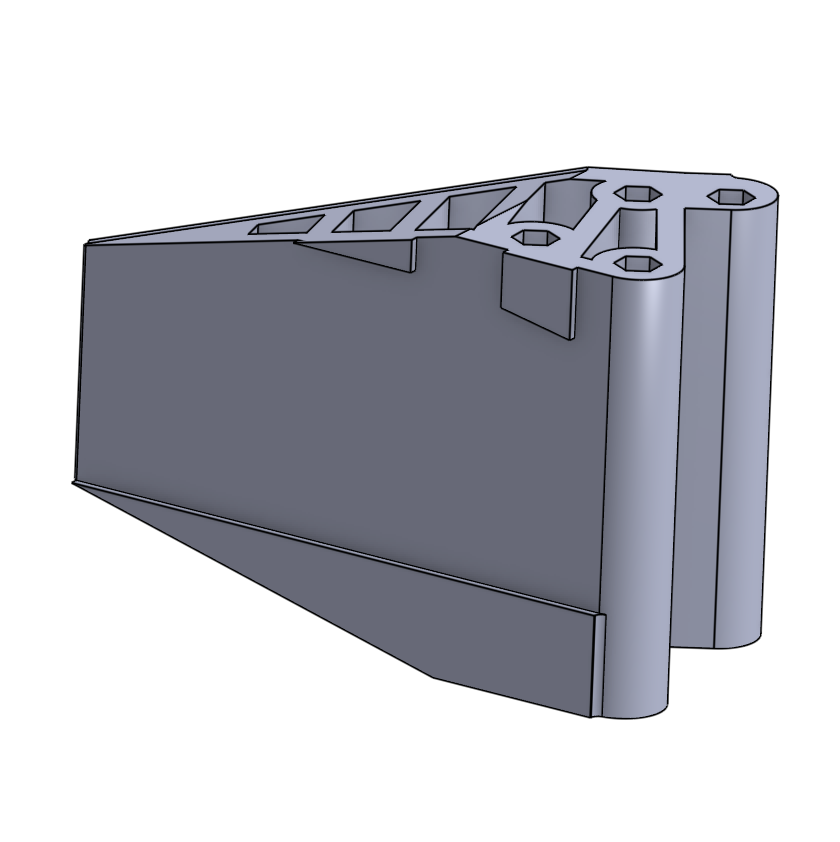}
            \caption{TPU gripper design with sensor slots}
            \label{fig:gripper_cad}
        \end{subfigure}
        \hspace{0.03\linewidth}
        \begin{subfigure}[t]{0.28\linewidth}
            \centering
            \includegraphics[width=\linewidth, height=6cm, keepaspectratio]{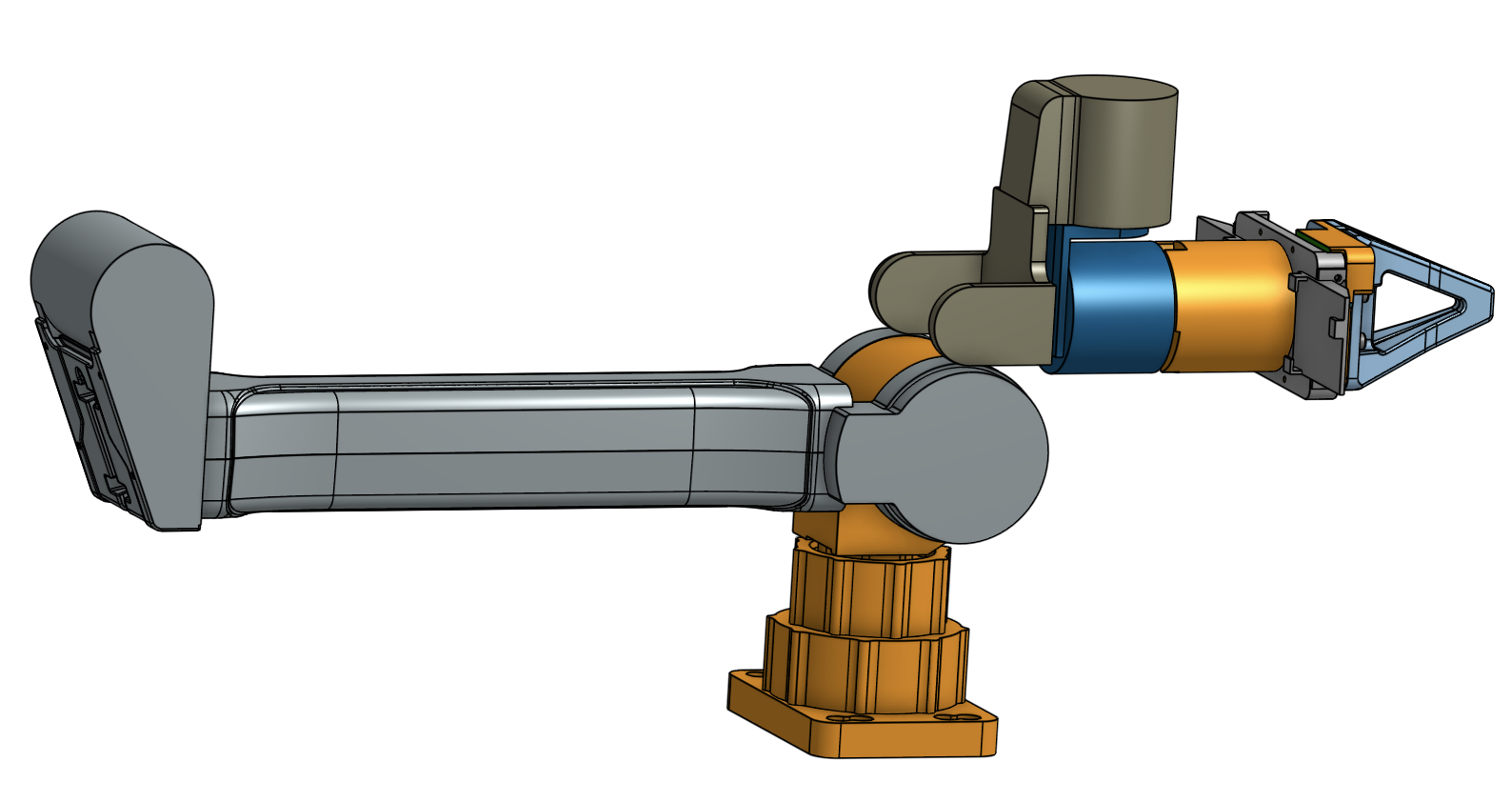}
            \caption{Single arm CAD with electronics mounts}
            \label{fig:whole_cad}
        \end{subfigure}
        \caption{Customized station design}
        \label{fig:hardware_cad}
    \end{figure}

\section{Training and Evaluation Details}
\label{app:training}

    \paragraph{Dataset construction.}
    All demonstrations are collected on the YAM station described in Sec.~\ref{sec:methodology}. RGB observations are recorded at approximately $30$~Hz and tactile observations at approximately $99$~Hz using a shared clock. The full-rate tactile stream is retained during recording, and each policy timestep is paired with the tactile measurement having the nearest timestamp. Each demonstration is stored as one episode. Training and validation sets are split by episode, with $10\%$ of episodes held out for validation using a fixed split by episode ID. Within each task, observation conditions use the same demonstration episodes and episode-level split. All experiments use random seed 42. ACT targets contain 19 consecutive future action steps; the data-loader stride changes the spacing between sampled training windows and does not subsample actions within the chunk.

    \begin{table}[htbp]
    \centering
    \small
    \caption{\textbf{Demonstration dataset sizes.}
    Total recorded episodes and successful episodes
    before the episode-level training/validation split.}
    \label{tab:demo_counts}
    \begin{tabular}{@{}lrr@{}}
    \toprule
    Task & Recorded & Successful \\
    \midrule
    T1: Block placement       & 267 & 229 \\
    T2: Block/sponge dataset  & 225 & 208 \\
    T3: Straw insertion       & 220 & 209 \\
    T4: Bag retrieval         & 235 & 203 \\
    T5: Plug insertion        & 231 & 202 \\
    T6: Whiteboard wiping     & 218 & 207 \\
    \midrule
    Total                     & 1396 & 1258 \\
    \bottomrule
    \end{tabular}
    \end{table}
    Table~\ref{tab:demo_counts} summarizes the recorded demonstration episodes and successful episodes for each task. Only successful demonstration episodes were retained for policy training before the episode-level training/validation split.

    \paragraph{Preprocessing.}
    RGB images are stored at $384{\times}384$ using JPEG quality 95 and processed at $224{\times}224$. During training, camera observations are resized, cropped to $224{\times}224$, normalized with ImageNet statistics, and augmented with color jitter $(0.2,0.2,0.2)$. Because the stored frames are square and the resize and crop sizes are identical, the nominal random crop does not introduce spatial augmentation in our setting. At evaluation, the corresponding deterministic preprocessing is used. Each tactile observation is a $32{\times}16$ grayscale map, replicated to three channels and resized directly to $224{\times}224$. Tactile inputs receive no crop or color augmentation and use the same ImageNet normalization.
    
    \paragraph{ACT architecture and optimization.}
    All experiments use ACT~\citep{act} with a shared ImageNet-pretrained ResNet-18~\citep{resnet} for camera and tactile views. The backbone uses \texttt{FrozenBatchNorm2d} and produces a $7{\times}7{\times}512$ feature map per view. The transformer uses model dimension 512, 8 attention heads, a 3200-dimensional feed-forward layer, four encoder layers, two decoder layers, four CVAE encoder layers, and a 32-dimensional latent variable. The model contains approximately 56.9M parameters. Table~\ref{tab:act_hparams} lists the optimization settings.
    
    \begin{table}[h]
    \centering
    \small
    \caption{\textbf{ACT training settings}}
    \label{tab:act_hparams}
    \begin{tabular}{@{}ll@{}}
    \toprule
    Setting & Value \\
    \midrule
    Action space & 7-D absolute joint positions \\
    Action chunk & 19 consecutive steps \\
    KL weight & 10.0 \\
    Batch size & 48 \\
    Optimizer & AdamW \\
    Learning rate & $10^{-5}$ \\
    Weight decay & $10^{-4}$ \\
    Gradient clipping & 1.0 global norm \\
    Warmup & 500 steps \\
    Schedule & cosine decay to 0 \\
    Training steps & 100,000 \\
    Precision & bf16 autocast \\
    Seed & 42 \\
    \bottomrule
    \end{tabular}
    \end{table}
    
    Training minimizes masked L1 action reconstruction plus the KL term,
    \[
    \mathcal{L}_{\mathrm{train}}
    =
    \mathcal{L}_{\mathrm{L1}}
    +
    10\,\mathcal{L}_{\mathrm{KL}}.
    \]
    
    \paragraph{Validation and checkpoints.}
    Validation is performed every 1,000 optimization steps on up to 50 batches from held-out episodes. In evaluation mode, ACT disables the CVAE branch, so the reported validation objective is masked L1 only rather than the full training objective. The training script saves \texttt{best.pt} whenever held-out L1 improves and numbered checkpoints every 10,000 steps. Reported evaluations use \texttt{best.pt} for all tasks except T5. For T5, we compared three checkpoints: \texttt{best.pt}, \texttt{last.pt}, and \texttt{80000.pt}, using 10 physical development rollouts per checkpoint for each of the four observation conditions. All four conditions followed the same checkpoint-selection procedure, and the 80,000-step checkpoint achieved higher observed success than the alternatives. We therefore selected \texttt{80000.pt} for the final T5 evaluations. All checkpoint-selection rollouts were conducted separately from and excluded from the 60 final evaluation rollouts per condition. Thus, T5 checkpoint selection was based on physical rollout performance rather than validation loss alone.
    
    \paragraph{Experimental conditions and rollout protocol.}
    The primary comparison is between \VO\ and \VT\ policies trained from the same task demonstrations. For T4--T6, this comparison is crossed with wrist-camera availability. When wrist images are disabled, scene and egocentric images remain available in both training and deployment.
    
    At deployment, ACT predicts a new 19-step action chunk at each policy timestep, and overlapping predictions are combined using temporal ensembling. Each rollout is limited to 900 control steps. Within each task, policies are evaluated using matched initial-condition sets, identical rollout counts, and the same success criterion. The reported evaluation comprises 30 rollouts per condition for T1, 25 each for T2a and T2b, and 60 for T3--T6, totaling 1000 rollouts. These are evaluation executions, not demonstration episodes.

    Table~\ref{tab:eval_allocation} lists the initial-condition allocation, which is shared across the observation conditions within each task. For T4, reported success rates pool the five positions and therefore reflect the nonuniform allocation of trials. Tables~\ref{tab:t1_t3_positions}--\ref{tab:t6_positions} give the corresponding outcome counts.

    \paragraph{Success criteria.}
    T1 requires successful placement in the target box. T2 requires successful grasping and placement in the box, with no sponge deformation for T2b. T3 requires the straw to be inserted through the cup-lid opening. T4 requires retrieval of the marker from the opaque bag; a trial in which the marker is subsequently lost before completion is counted as a failure. T5 requires the plug to be inserted in the socket. T6 requires at least $50\%$ of the marked whiteboard region to be erased. In T6, successful eraser pickup is also recorded separately to distinguish pickup failures from failures after pickup.
    
    \begin{table}[htbp]
      \centering
      \small
      \caption{\textbf{Evaluation allocation per observation condition.}}
      \label{tab:eval_allocation}
      \begin{tabular}{@{}lp{0.52\linewidth}lr@{}}
        \toprule
        Task & Initial conditions & Trials per position & Total \\
        \midrule
        T1 & Middle, top left, top right, bottom left, bottom right, far right in front of the box & 5 each & 30 \\
        T2a/b & Middle, top left, top right, bottom left, bottom right & 5 each & 25 each \\
        T3 & Middle, top left, top right, bottom left, bottom right & 12 each & 60 \\
        T4 & Far left, left, middle, right, far right & 10, 5, 5, 20, 20 & 60 \\
        T5 & Left, right, middle, top, bottom & 12 each & 60 \\
        T6 & Top, center, bottom & 20 each & 60 \\
        \bottomrule
      \end{tabular}
    \end{table}

\clearpage
\section{Posterior Success Probabilities and Pairwise Tests}
\label{app:posteriors}

\begin{figure}[h]
    \centering
    \includegraphics[width=\linewidth]{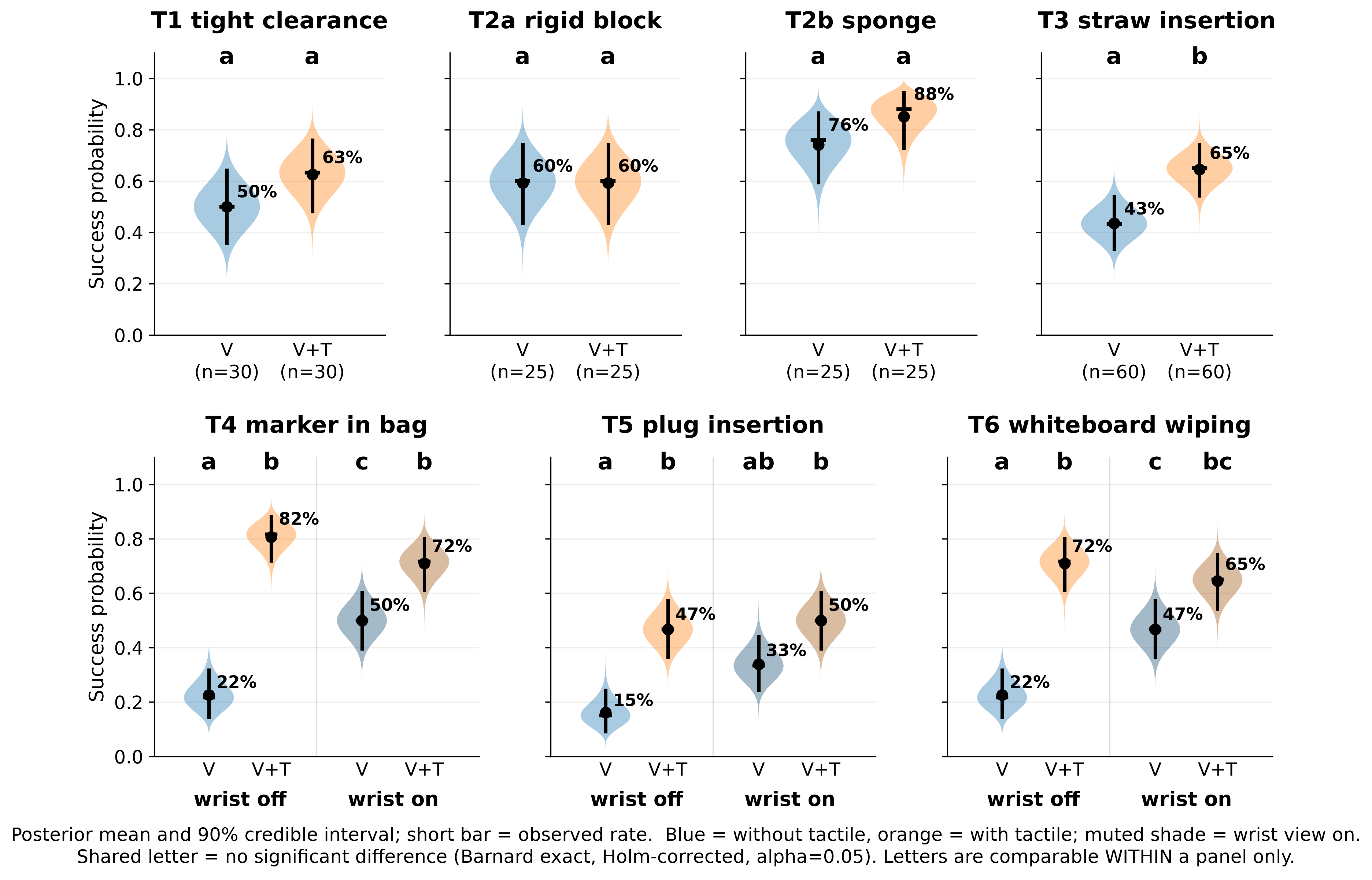}
    \caption{\textbf{Posterior success probability for every task, with compact letter displays.} Beta posteriors under a uniform prior. Top row: the wrist-on tasks (T1--T3), two conditions each. Bottom row: the ablated tasks (T4--T6), four conditions each. Letters come from Holm-corrected pairwise Barnard tests computed within each panel; conditions sharing a letter could not be distinguished at $\alpha{=}0.05$. Letters are not comparable across panels. In T4 and T6 the two \VT\ conditions share a letter while the two \VO\ conditions do not---no significant wrist-on versus wrist-off difference was detected for \VT\ in these tasks.}
    \label{fig:violin}
    \end{figure}

    Table~\ref{tab:results} reports success-rate point estimates. For $s$ successes in $n$ rollouts, a uniform $\operatorname{Beta}(1,1)$ prior yields the posterior
    \[
      p \mid s,n \sim \operatorname{Beta}(s+1,n-s+1).
    \]
    Figure~\ref{fig:violin} shows the condition-wise posteriors together with compact letter displays~\citep{lbm_paper}. Pairwise comparisons use Barnard tests, with Holm correction applied within each task panel. The four-condition panels for T4--T6 each contain six pairwise comparisons; each two-condition panel contains one comparison. Conditions sharing a letter are not distinguishable at the corrected $\alpha=0.05$ level. Failure to reject a difference does not establish equivalence.
    
    The tests and posterior intervals concern the evaluated policies and rollout allocation. They do not account for training-seed variation or checkpoint-selection uncertainty. The additional position-level comparisons in the main text and Appendix~\ref{app:counts} are descriptive post hoc analyses; no subgroup significance claims are made.

    \begin{figure}[H]
    \centering
    \includegraphics[width=0.75\linewidth]{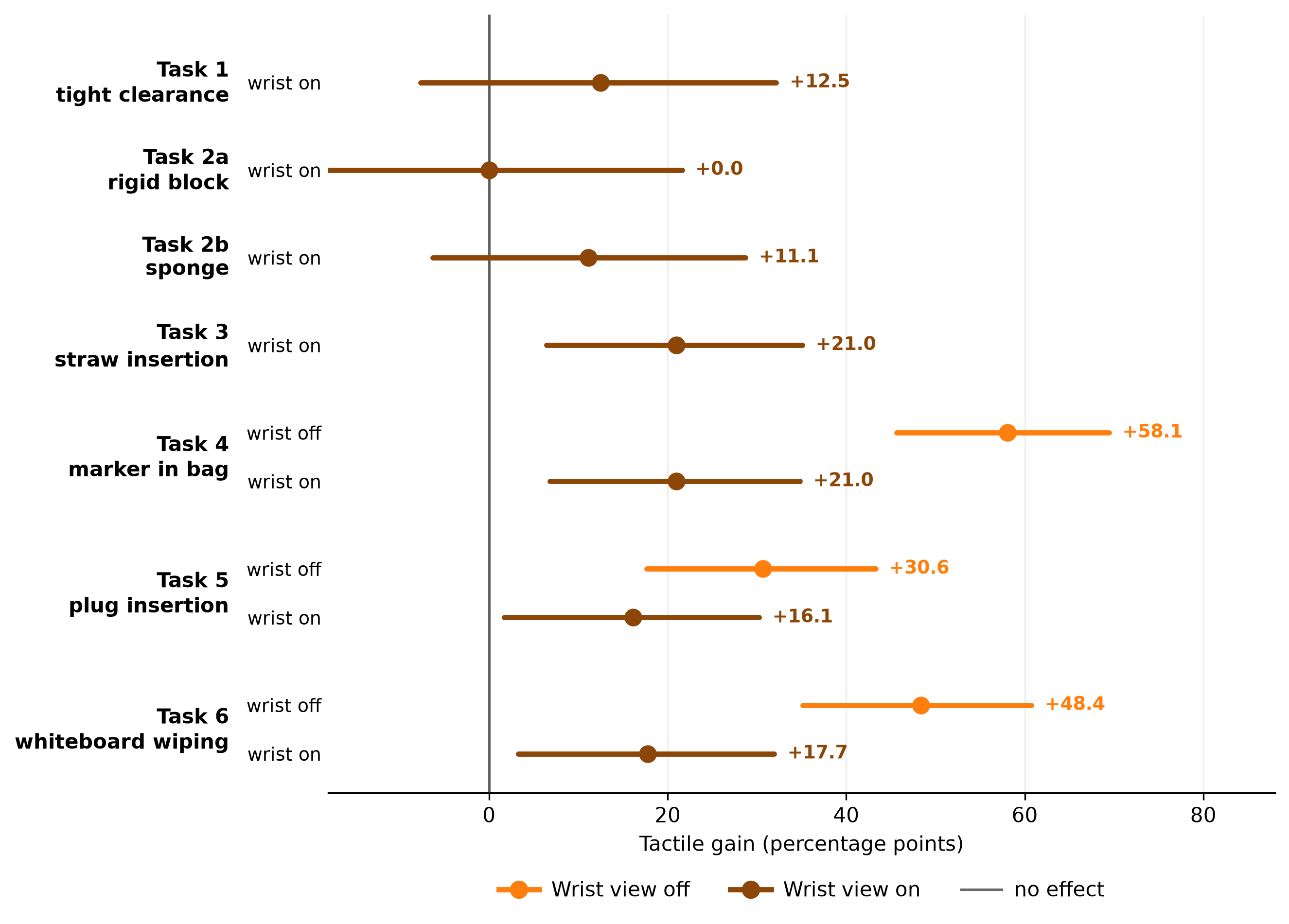}
    \caption{\textbf{Tactile gain by task and wrist-camera availability.}
    Posterior mean and $90\%$ credible interval for the difference in success probability between \VT\ and \VO.}
    \label{fig:gain}
    \end{figure}
    
    \clearpage
    
    \section{Evaluation Outcomes by Initial Condition}
    \label{app:counts}
    
    The following tables provide the recorded numerators and denominators underlying Table~\ref{tab:results}. T2a and T2b are reported as separate evaluation variants. All counts are evaluation outcomes. The breakdowns support the descriptive analyses in Sec.~\ref{sec:wrist} and Sec.~\ref{sec:substitute}; they do not constitute additional training runs or checkpoint evaluations.
    
    \begin{table}[htbp]
      \centering
      \small
      \caption{\textbf{T1--T3 success counts by initial condition.}
      Wrist images are enabled. Each entry is successes/rollouts.}
      \label{tab:t1_t3_positions}
      \begin{tabular}{@{}llcc@{}}
        \toprule
        Task & Initial condition & \VO & \VT \\
        \midrule
        T1 & Middle & 4/5 & 5/5 \\
           & Top left & 2/5 & 3/5 \\
           & Top right & 0/5 & 2/5 \\
           & Bottom left & 1/5 & 0/5 \\
           & Bottom right & 5/5 & 5/5 \\
           & Far right in front of the box & 3/5 & 4/5 \\
           & Total & 15/30 & 19/30 \\
        \midrule
        T2a & Middle & 2/5 & 2/5 \\
            & Top left & 3/5 & 3/5 \\
            & Top right & 4/5 & 3/5 \\
            & Bottom left & 3/5 & 4/5 \\
            & Bottom right & 3/5 & 3/5 \\
            & Total & 15/25 & 15/25 \\
        \midrule
        T2b & Middle & 4/5 & 4/5 \\
            & Top left & 3/5 & 4/5 \\
            & Top right & 4/5 & 5/5 \\
            & Bottom left & 4/5 & 4/5 \\
            & Bottom right & 4/5 & 5/5 \\
            & Total & 19/25 & 22/25 \\
        \midrule
        T3 & Middle & 5/12 & 6/12 \\
           & Top left & 8/12 & 8/12 \\
           & Top right & 8/12 & 9/12 \\
           & Bottom left & 2/12 & 7/12 \\
           & Bottom right & 3/12 & 9/12 \\
           & Total & 26/60 & 39/60 \\
        \bottomrule
      \end{tabular}
    \end{table}
    
    \begin{table}[htbp]
      \centering
      \small
      \caption{\textbf{T4 retrieval success by initial position.}
      Each entry is successes/rollouts. The aggregate \VT\ wrist-on decrease is concentrated in the far-right group; the other positions yield $34/40$ successes in both \VT\ configurations.}
      \label{tab:t4_positions}
      \begin{tabular}{@{}lcccc@{}}
        \toprule
         & \multicolumn{2}{c}{Wrist off} & \multicolumn{2}{c}{Wrist on} \\
        \cmidrule(lr){2-3}\cmidrule(l){4-5}
        Initial position & \VO & \VT & \VO & \VT \\
        \midrule
        Far left & 3/10 & 9/10 & 5/10 & 8/10 \\
        Left & 4/5 & 5/5 & 5/5 & 4/5 \\
        Middle & 2/5 & 4/5 & 5/5 & 5/5 \\
        Right & 4/20 & 16/20 & 11/20 & 17/20 \\
        Far right & 0/20 & 15/20 & 4/20 & 9/20 \\
        \midrule
        Total & 13/60 & 49/60 & 30/60 & 43/60 \\
        \bottomrule
      \end{tabular}
    \end{table}
    
    \begin{table}[htbp]
      \centering
      \small
      \caption{\textbf{T5 insertion success by initial position.}
      Each entry is successes/rollouts.}
      \label{tab:t5_positions}
      \begin{tabular}{@{}lcccc@{}}
        \toprule
         & \multicolumn{2}{c}{Wrist off} & \multicolumn{2}{c}{Wrist on} \\
        \cmidrule(lr){2-3}\cmidrule(l){4-5}
        Initial position & \VO & \VT & \VO & \VT \\
        \midrule
        Left & 0/12 & 8/12 & 3/12 & 7/12 \\
        Right & 1/12 & 3/12 & 5/12 & 3/12 \\
        Middle & 3/12 & 8/12 & 6/12 & 7/12 \\
        Top & 4/12 & 4/12 & 4/12 & 6/12 \\
        Bottom & 1/12 & 5/12 & 2/12 & 7/12 \\
        \midrule
        Total & 9/60 & 28/60 & 20/60 & 30/60 \\
        \bottomrule
      \end{tabular}
    \end{table}
    
    \begin{table}[htbp]
      \centering
      \small
      \caption{\textbf{T6 pickup and task success by initial position.}
      Each policy entry is $(P,S)$, where $P$ is the number of successful pickups and $S$ is the number of successful complete trials. The trial count $n$ applies to every policy in that row.}
      \label{tab:t6_positions}
      \begin{tabular}{@{}lrcccc@{}}
        \toprule
         & & \multicolumn{2}{c}{Wrist off} & \multicolumn{2}{c}{Wrist on} \\
        \cmidrule(lr){3-4}\cmidrule(l){5-6}
        Initial position & $n$ & \VO & \VT & \VO & \VT \\
        \midrule
        Top & 20 & $(5,4)$ & $(17,13)$ & $(8,7)$ & $(18,15)$ \\
        Center & 20 & $(15,6)$ & $(16,15)$ & $(17,15)$ & $(15,15)$ \\
        Bottom & 20 & $(3,3)$ & $(15,15)$ & $(6,6)$ & $(9,9)$ \\
        \midrule
        Total & 60 & $(23,13)$ & $(48,43)$ & $(31,28)$ & $(42,39)$ \\
        \bottomrule
      \end{tabular}
    \end{table}

\end{document}